\documentclass[conference]{IEEEtran}
\IEEEoverridecommandlockouts
\usepackage{amsmath,amssymb,amsthm}
\usepackage{graphicx}
\usepackage{booktabs}
\usepackage{cite}
\usepackage{algorithm}
\usepackage{algpseudocode}
\usepackage{xcolor}
\usepackage[hidelinks]{hyperref}
\newtheorem{proposition}{Proposition}
\newtheorem{remark}{Remark}
\newcommand{\trackOneMin}{64}
\newcommand{\trackTwoMin}{73}
\newcommand{\insErrMin}{92}
\newcommand{\insErrMax}{2203}
\newcommand{\mapErrOffOne}{3.2}
\newcommand{\mapErrOffTwo}{3.4}
\newcommand{\mapErrSurOne}{21.5}
\newcommand{\mapErrSurTwo}{19.9}
\newcommand{\nCondSig}{9}
\newcommand{\nCond}{12}
\newcommand{\ratioMin}{0.55}
\newcommand{\ratioMax}{0.92}
\newcommand{\rOneRatioA}{0.91}
\newcommand{\rOneRatioB}{0.92}
\newcommand{\rOnePB}{0.08}
\newcommand{\rThreeWholeLo}{20}
\newcommand{\rThreeWholeHi}{27}
\newcommand{\rThreeLastLo}{32}
\newcommand{\rThreeLastHi}{40}
\newcommand{\rFourWholeA}{45}
\newcommand{\rFourLastA}{79}
\newcommand{\rFourWinsA}{28}
\newcommand{\rFourWholeB}{30}
\newcommand{\rFourLastB}{38}
\newcommand{\rFourMedTheoryA}{43}
\newcommand{\rFourMedBestA}{199}
\newcommand{\divTheoryMin}{1}
\newcommand{\divTheoryMax}{2}
\newcommand{\divBaseMin}{1}
\newcommand{\divBaseMax}{2}
\newcommand{\infoBetterCount}{3}
\newcommand{\degrSig}{4}
\newcommand{\degrLo}{10}
\newcommand{\degrHi}{25}
\newcommand{\swMidLo}{30}
\newcommand{\swMidHi}{44}
\newcommand{\swMidLastHi}{54}
\newcommand{\swMidPmax}{0.012}
\newcommand{\swHighRatio}{0.96}
\newcommand{\swLowRatio}{0.93}
\newcommand{\swHighP}{0.43}
\newcommand{\swLowP}{0.47}

\newcommand{\trSdStart}{96}
\newcommand{\trSdEnd}{3.7}
\newcommand{\trTauStart}{51}
\newcommand{\trTauEnd}{733}
\newcommand{\trErrEnd}{32}
\newcommand{\planJointLo}{23}
\newcommand{\planJointHi}{46}

\newcommand{\planLightLo}{27}
\newcommand{\planLightHi}{43}
\newcommand{\planOvLight}{9}
\newcommand{\planOvFull}{29}
\newcommand{\planDivMax}{1}
\newcommand{\hwTtwo}{202}
\newcommand{\hwBeta}{1.0}
\newcommand{\hwAzero}{0.47}
\newcommand{\hwJob}{dbvtfddhdvas73bil3m0}
\newcommand{\hwAdaptWorst}{1.3}
\newcommand{\hwFixedBest}{5.4}
\newcommand{\hwFixedBestTau}{7}
\newcommand{\hwFixedWorst}{48}
\newcommand{\abRthreeA}{1.00}
\newcommand{\abRthreeB}{0.98}
\newcommand{\abRthreePmin}{0.5}
\newcommand{\abRfourLo}{37}
\newcommand{\abRfourHi}{42}
\newcommand{\abRfourPmax}{3\times10^{-4}}
\newcommand{\abBPfourA}{0.91}
\newcommand{\abBPfourB}{1.14}
\newcommand{\abFPfourLo}{20}
\newcommand{\abFPfourHi}{27}
\newcommand{\abBPthreeLo}{19}
\newcommand{\abBPthreeHi}{24}
\newcommand{\abLadMin}{0.61}
\newcommand{\abLadMax}{0.94}
\newcommand{\abRoneA}{0.95}
\newcommand{\abRoneB}{0.87}
\newcommand{\abFPmeanRA}{625}
\newcommand{\abThMeanRA}{805}

\newcommand{\abFPmeanRB}{498}
\newcommand{\abThMeanRB}{753}

\newcommand{\holmSig}{6}
\newcommand{\winWins}{63}
\newcommand{\winN}{72}
\newcommand{\plBProuteLo}{7}
\newcommand{\plBProuteHi}{18}
\newcommand{\plBProuteSig}{2}
\newcommand{\plJointBPLo}{15}
\newcommand{\plJointBPHi}{40}
\newcommand{\plJointBPPmax}{0.02}

\newcommand{\plMidJointInfo}{28}
\newcommand{\plLostJointInfo}{8}

\begin{document}

\title{Belief-Driven Quantum Magnetometer Control for GPS-Denied Navigation}

\author{\IEEEauthorblockN{Murtuza Lemonwala and Laxima Niure Kandel}
\IEEEauthorblockA{Department of Electrical Engineering and Computer Science\\
Embry-Riddle Aeronautical University, Daytona Beach, FL, USA\\
lemonwalaresearch@gmail.com, niurekal@erau.edu}}

\maketitle

\begin{abstract}
Autonomous vehicles increasingly need to navigate without satellite positioning. Magnetic anomaly navigation (MagNav) with quantum magnetometers offers a passive, unjammable fix, but the low size-weight-and-power (SWaP) sensors that small drones can carry are weak, and they are operated open loop at a fixed setting. Interferometric (Ramsey) magnetometers face a trade-off: long interrogation times are more sensitive but become ambiguous when the expected field is uncertain. The navigator's own position belief, pushed through the anomaly map, predicts that field. We close this loop. The filter's predictive field distribution sets the Ramsey phase and interrogation time at every epoch, through a closed-form rule that moves between an ambiguity limit and a decoherence limit. We evaluate it on real U.S. Air Force flight data (real tracks, inertial drift, measured fields, and a survey map tested 18 months after it was flown), with only the Ramsey sensor modelled. Against fixed-setting baselines, 63 of 72 independent flight windows favour belief-driven sensing ($p=2\times10^{-11}$). Error falls by 20--27\% for low-SNR sensors and 30--44\% for mid-contrast sensors. An ablation separates the two ingredients. Phase alignment drives the gain in steady flight, and the rule matches the best fixed setting without the hindsight tuning that setting requires. After a lost start, interrogation-time adaptation is essential, cutting median error by \abRfourLo--\abRfourHi\% even against hindsight-tuned phase-aligned settings. Gains are small for high-quality sensors with accurate maps. On an IBM superconducting qubit, the rule's interrogation time stays within $1.3\times$ of the empirical optimum. Coupled with information-aware routing, belief-driven sensing lowers mission error by \plJointBPLo--\plJointBPHi\%.

\end{abstract}


\section{Introduction}
Satellite navigation is easy to jam and spoof, and many autonomous vehicles must now operate without it. Small uncrewed aircraft systems (sUAS) are among the most exposed: they rely on low-cost inertial sensors that drift quickly, and they cannot carry the size, weight and power (SWaP) of the navigation suites used on large platforms. They need an alternative position fix that is passive, cheap and compact.

Magnetic anomaly navigation (MagNav) bounds inertial drift by matching measurements of the Earth's crustal field against an anomaly map~\cite{canciani2017,gnadt2023signal}. It is passive and practically unjammable. Quantum magnetometers make MagNav practical on moving platforms, and it has matured quickly:
\begin{itemize}
\item airborne trials report positioning better than a strategic-grade inertial system~\cite{qctrl2025};
\item real-time MagNav has flown on a military transport aircraft~\cite{afmagnav2023};
\item quantum-enabled MagNav has been flight-tested on a Group~3 sUAS~\cite{lumberjack2026}.
\end{itemize}
The open question for drones is SWaP. The compact sensors a small aircraft can carry have far weaker readout than laboratory or airliner-class instruments.

In fielded systems the quantum sensor is operated \emph{open loop}: its operating point is fixed in advance, and the navigation filter consumes its output. An interferometric magnetometer, however, faces a trade-off that depends on what the navigator already knows. In a Ramsey sequence~\cite{ramsey1950,degen2017} the sensor accumulates a phase $2\pi\gamma\tau B$ over an interrogation time $\tau$. A longer $\tau$ makes each shot more sensitive. It also shrinks the range of fields that can be read without $2\pi$ ambiguity, and leaves fewer shots per second once overhead is included. Adaptive phase estimation manages this trade-off for a \emph{stationary} sensor and a field-accuracy objective~\cite{said2011,bonato2016,ma2025adaptive,cooke2025rl,le2025distributed}. A navigating drone is in a different position: its own belief, pushed through the map, already predicts the field it is about to measure. That prediction is broad after a jamming-induced loss of position and narrow once the vehicle is localized.

\textbf{This paper closes that loop.} The navigation filter's predictive belief sets the Ramsey phase and interrogation time at every epoch. We evaluate the idea on real airborne data and identify which sensors and situations benefit.

\textbf{Contributions.}
\begin{enumerate}
\item \emph{Belief-driven sensing.} The map-matching filter's predictive field distribution sets two things every epoch. It aligns the readout phase to the predicted field. It also chooses the interrogation time, so the map acts as side information that resolves phase ambiguity, much as a classical accelerometer does in hybrid atom-interferometer inertial sensors~\cite{cheiney2018}.
\item \emph{A closed-form design rule.} Under a Gaussian predictive belief we derive the interrogation time that maximizes visibility-weighted Fisher information (Proposition~\ref{prop:tau}). It moves between a decoherence limit and an \emph{ambiguity limit} set by the navigator's own uncertainty. We also give an expected-information-gain rule for multi-modal beliefs.
\item \emph{Evaluation on real flight data, with ablations.} Using the DAF-MIT MagNav flights~\cite{dafmitdata,gnadt2023signal}, we compare against hindsight-tuned fixed settings, phase-aligned fixed settings, and dynamic-range-safe ladders~\cite{kimmel2015}. We test across sensor classes, a lost start, and a degraded map, and we separate the contributions of phase alignment and interrogation-time adaptation.
\item \emph{Hardware check.} Ramsey circuits with real delays on an IBM superconducting qubit confirm the trade-off and the rule's choice of interrogation time.
\item \emph{Joint route and sensor control.} An information-aware route planner scores routes with the sensor schedule that will actually be used along them.
\end{enumerate}

\section{Related Work}
\emph{Quantum-sensor navigation.} Lellouch and Holynski integrate a high-fidelity atom-interferometer gradiometer model with a particle-filter map matcher~\cite{lellouch2025}. Their sensor parameters are fixed per scenario, and they conclude that navigation filtering and quantum-sensor optimization should be considered jointly. The same group and industry partners have since extended this simulation approach to quantum inertial sensors in realistic navigation contexts~\cite{towlson2026}. Michael \emph{et al.}\ built a simulation platform for NV-diamond magnetometers on UAV and LEO platforms~\cite{michael2026}. They examine how configuration choices such as wide versus narrow dynamic range affect anomaly detection and navigation. Their focus is detecting sensor anomalies, not using the navigation belief to set the operating point. Le \emph{et al.}\ use reinforcement learning to adapt NV Ramsey settings within a one-shot geolocalization~\cite{le2025distributed}; there is no navigation filter in that loop. Field trials with proprietary quantum magnetometers~\cite{qctrl2025} focus on denoising and map matching.

\emph{Adaptive quantum magnetometry.} Bayesian and reinforcement-learning protocols choose sensing times and phases to estimate a static field with both high sensitivity and high dynamic range~\cite{said2011,bonato2016,ma2025adaptive,cooke2025rl}. Their objective is field accuracy for a sensor at rest. Our objective is position information for a moving sensor whose prior comes from a map.

\emph{Information-aware MagNav.} Active localization chooses actions to reduce posterior uncertainty~\cite{fox1998active}. For MagNav, information-aware motion planners steer toward informative terrain~\cite{penumarti2024realtime,penumarti2024global}, and optimal-control trajectories minimize the posterior Cram\'er--Rao bound~\cite{kang2026optimal}. Deep reinforcement learning has also been applied to geomagnetic navigation~\cite{bai2024geomag}. These works treat the sensor as fixed. We adapt the sensor, and then plan the route and the sensor together.

\section{Problem Formulation}\label{sec:problem}
\subsection{Navigation filter}
A vehicle with horizontal position $\mathbf{x}_t$ receives inertial displacements $\Delta\mathbf{x}^{\mathrm{INS}}_t$ corrupted by a slowly varying velocity bias $\mathbf{b}_t$. A particle filter~\cite{arulampalam2002} carries $N_p$ hypotheses $(\mathbf{x}^{(i)}_t,\mathbf{b}^{(i)}_t)$ with weights $w^{(i)}_t$:
\begin{equation}
\mathbf{b}^{(i)}_{t}=\mathbf{b}^{(i)}_{t-1}+\boldsymbol\eta_b,\qquad
\mathbf{x}^{(i)}_{t}=\mathbf{x}^{(i)}_{t-1}+\Delta\mathbf{x}^{\mathrm{INS}}_t-\mathbf{b}^{(i)}_{t}+\boldsymbol\eta_x .
\end{equation}
Each hypothesis predicts the anomaly field $m(\mathbf{x}^{(i)}_t)$ from the map $m(\cdot)$. The true field differs from the map by a map error $e_t\sim\mathcal N(0,\sigma_m^2)$.

\subsection{Ramsey measurement model}
In every epoch of duration $T_b$ the magnetometer runs $n(\tau)=\lfloor T_b/(\tau+t_0)\rfloor$ Ramsey shots with interrogation time $\tau$, readout phase $\phi$, and per-shot overhead $t_0$. The bright-state probability is
\begin{equation}\label{eq:ramsey}
p(B;\tau,\phi)=\tfrac12\!\left[1+C\,e^{-(\tau/T_2^*)^2}\cos\!\left(2\pi\gamma\tau B+\phi\right)\right],
\end{equation}
where $C$ is the effective readout contrast, $T_2^*$ the dephasing time, and $\gamma=28$\,Hz/nT. The bright count is $k\sim\mathrm{Bin}(n,p)$. The filter updates each particle with the exact binomial likelihood, marginalized over the map error by Gauss--Hermite quadrature:
\begin{equation}
w^{(i)}_t\propto w^{(i)}_{t-1}\!\int \mathrm{Bin}\!\left(k\,\middle|\,n,p\big(m(\mathbf{x}^{(i)}_t)+e\big)\right)\mathcal N(e;0,\sigma_m^2)\,de .
\end{equation}
Because the likelihood is periodic in $B$, the filter itself performs map-aided phase unwrapping. Distant hypotheses that alias onto the same fringe are not separated by that measurement, but are separated later as the vehicle moves.

\subsection{Sensing policy}
A sensing policy chooses $(\tau_t,\phi_t)$ before each epoch from information available at that time. All policies below share the same filter, the same likelihood, and the same per-epoch time budget. They differ \emph{only} in how they spend the budget.

\section{Belief-Driven Sensing}\label{sec:method}
\subsection{Predictive field belief}
Before each epoch, the particles define a predictive distribution of the field to be measured, with mean and variance
\begin{equation}
\mu_t=\sum_i w^{(i)}m(\mathbf{x}^{(i)}_t),\quad
\sigma_t^2=\sum_i w^{(i)}\big(m(\mathbf{x}^{(i)}_t)-\mu_t\big)^2+\sigma_m^2 .
\end{equation}
The phase is set so that the predicted field sits at the steepest point of the fringe, $\phi_t=\pi/2-2\pi\gamma\tau_t\mu_t$.

\subsection{Closed-form interrogation time}
If the predictive distribution is approximately Gaussian, the fringe in \eqref{eq:ramsey}, averaged over it, has visibility reduced by $\mathbb E[e^{i2\pi\gamma\tau(B-\mu_t)}]=e^{-(2\pi\gamma\tau\sigma_t)^2/2}$. The Fisher information about $B$ per shot at quadrature is $(2\pi\gamma\tau C_{\mathrm{eff}})^2$. Weighting by the predictive visibility gives the epoch score
\begin{equation}\label{eq:score}
J(\tau)=n(\tau)\,\tau^2\exp\!\Big[-\tfrac{2\tau^2}{T_2^{*2}}-(2\pi\gamma\sigma_t)^2\tau^2\Big].
\end{equation}
A measurement taken over flat terrain still carries little position information. That factor (the squared map gradient) does not depend on $\tau$, so it does not change the choice of $\tau$.

\begin{proposition}\label{prop:tau}
For negligible overhead ($t_0\ll\tau$, so $n\propto1/\tau$), $J$ is maximized at
\begin{equation}\label{eq:taustar}
\tau^\star=\left(\frac{4}{T_2^{*2}}+2\,(2\pi\gamma\sigma_t)^2\right)^{-1/2}.
\end{equation}
\end{proposition}
\begin{proof}
With $n\propto1/\tau$, $\ln J=\ln\tau-a\tau^2+\text{const}$ with $a=2/T_2^{*2}+(2\pi\gamma\sigma_t)^2$. Setting $d\ln J/d\tau=1/\tau-2a\tau=0$ gives $\tau^\star=(2a)^{-1/2}$, and the second derivative is negative.
\end{proof}

\begin{remark}
Equation~\eqref{eq:taustar} has two regimes. When the vehicle is well localized ($2\pi\gamma\sigma_t T_2^*\ll1$), $\tau^\star\to T_2^*/2$: the sensor is limited by \emph{decoherence}. When the vehicle is uncertain, $\tau^\star\to1/(\sqrt2\,2\pi\gamma\sigma_t)$: the sensor is limited by \emph{ambiguity}, at the scale set by the navigator's own uncertainty. A heuristic $\tau=\kappa/(2\pi\gamma\sigma_t)$ is therefore optimal at $\kappa=1/\sqrt2$. With overhead, we maximize \eqref{eq:score} numerically over a grid of $\tau$, which takes microseconds. We call this policy \textsc{Theory}.
\end{remark}

\subsection{Expected-information-gain rule}
When the vehicle is lost, the predictive belief is multi-modal and the Gaussian visibility argument is crude. The \textsc{Info} policy instead picks the $\tau$ that maximizes the mutual information between the epoch's outcome and the particle hypothesis, which is the classical active-localization criterion~\cite{fox1998active}. It is estimated by Monte Carlo over $256$ weight-resampled particles. For each candidate $\tau$ and each hypothesis we form the map-error-marginalized outcome mean and variance (Gaussian approximation of $k/n$), draw one outcome, and average $\log p(y\,|\,i)-\log\frac1{256}\sum_j p(y\,|\,j)$.

\subsection{Joint route and sensor control}\label{sec:plan}
Belief-driven sensing changes \emph{which routes are informative}. A gradient-rich region is valuable only if the sensor can read it without ambiguity. A fixed long-$\tau$ sensor loses fringe visibility there whenever the vehicle is uncertain, while an adaptive sensor shortens $\tau$ and keeps the information.

Every 10\,s the planner scores candidate headings: offsets of $\{0,\pm25^\circ\}$ (\emph{Info-light}) or $\{0,\pm25^\circ,\pm50^\circ\}$ (\emph{Info}) from the bearing to the goal, with a turn-rate limit. Candidates that leave the surveyed area (minus a 600\,m margin) are rejected. For each candidate it rolls the particle covariance $\mathbf P$ forward over the next 10\,s along the leg and adds, at every second, the map-gradient information
\begin{equation}
\mathbf P^{-1}\leftarrow\mathbf P^{-1}+\frac{\nabla m\,\nabla m^{\top}}{F_B(\tau,\sigma)^{-1}+\sigma_m^2},\qquad \sigma^2=\nabla m^{\top}\mathbf P\,\nabla m+\sigma_m^2,
\end{equation}
where $F_B$ is the visibility-weighted Fisher information of \eqref{eq:score}. The planner picks the heading that minimizes $\mathrm{tr}\,\mathbf P$ (A-optimality). The only difference between the fixed-sensor and the joint planner is which $\tau$ enters $F_B$: the fixed $\tau$, or the $\tau^\star(\sigma)$ of Proposition~\ref{prop:tau} that the vehicle will actually use there.

\section{Experimental Setup}\label{sec:setup}
\subsection{Flight data and maps}
We use the DAF-MIT open MagNav flight data~\cite{dafmitdata,gnadt2023signal}. These were collected by Sander Geophysics Ltd.\ near Ottawa, Canada, in a Cessna Grand Caravan carrying optically pumped cesium magnetometers, fluxgates, a navigation-grade INS, and GPS. Ground truth is the GPS track. The \emph{true field} is the compensated, IGRF- and diurnal-corrected anomaly from the tail-stinger magnetometer (\texttt{mag\_1\_igrf}). The inertial input is the aircraft's own INS solution, so drift is real rather than simulated. It is converted to displacements in UTM zone~18N. The filter uses only INS displacements, so what matters is the INS drift rate. For reference, the absolute INS error at the start of the test windows ranges from \insErrMin\,m to \insErrMax\,m. Each window starts from a Gaussian position prior with standard deviation $\sigma_0$ (300\,m, or 1500\,m for the lost start), centred at the true position plus a random offset drawn from the same distribution.

We test on two flights over the Perth mini-survey area at 800\,m altitude:
\begin{itemize}
\item Flight~1007, line~1007.02 (free flight, July 2020, \trackOneMin\,min);
\item Flight~2016, line~2005.36 (a re-flight of the same path, January 2022, \trackTwoMin\,min), about 18 months later.
\end{itemize}
Data are used at 1\,Hz epochs.

\emph{Official map.} The MagNav.jl compilation of the Perth survey at 800\,m. Its error against the measured field along the test tracks, after removing a constant level offset, is \mapErrOffOne\,nT (Flight~1007) and \mapErrOffTwo\,nT (Flight~2016), standard deviation. We set $\sigma_m=3.5$\,nT.

\emph{Degraded map.} As a stress test, we built our own map by linear interpolation of the available survey lines. The public data withhold four survey lines as hold-out data. The resulting map error is \mapErrSurOne\,nT and \mapErrSurTwo\,nT, correlated over about 10\,s ($\approx$700\,m), and we set $\sigma_m=20$\,nT.

\subsection{Sensor classes}
The Ramsey sensor is modelled, because the dataset records continuous optically-pumped magnetometers rather than Ramsey shots. We use $T_2^*=2$\,ms and three generic classes, each with a per-second sensing budget $T_b$:
\begin{itemize}
\item \textbf{R1, ensemble}: $C=0.6$, $t_0=2$\,ms, $T_b=0.2$\,s;
\item \textbf{R3, low-SNR}: $C=0.03$, $t_0=0.05$\,ms, $T_b=0.02$\,s, a stand-in for weak-readout, low size-weight-and-power sensors such as single-spin or small-ensemble NV devices~\cite{barry2020};
\item \textbf{R4, lost start}: $C=0.1$, $t_0=0.2$\,ms, $T_b=0.05$\,s, with the initial position uncertainty raised from 300\,m to 1500\,m.
\end{itemize}
A contrast sweep ($C$ from 0.02 to 0.6, $t_0=0.2$\,ms, $T_b=0.05$\,s) maps where adaptivity matters.

\subsection{Baselines and protocol}
\begin{itemize}
\item \textsc{Fixed}($\tau$), with $\tau\in\{20,100,300,1000\}\,\mu$s and the phase at quadrature of the reference field. We report the best fixed $\tau$ chosen \emph{with hindsight} for each case, which is an oracle-tuned baseline.
\item \textsc{Ladder}: a dynamic-range-safe geometric ladder of $\tau$ in the style of robust phase estimation~\cite{kimmel2015}, with equal time per rung. \textsc{Ladder2} reads both quadratures on every rung.
\item \emph{Ablations}:
  \begin{itemize}
  \item \textsc{Fixed}($\tau$)+BP: a fixed $\tau$ with the phase aligned to the predicted field ($\phi_t$ of Sec.~\ref{sec:method}), again best $\tau$ with hindsight;
  \item \textsc{Ladder}+BP: the ladder with the same phase alignment;
  \item $\tau$\textsc{-only}: the Proposition-1 $\tau$ with the fixed reference phase.
  \end{itemize}
\item \textsc{Heur}($\kappa$): the ambiguity heuristic $\tau=\kappa/(2\pi\gamma\sigma_t)$, with $\kappa\in\{0.5,1\}$, capped at the decoherence optimum.
\item The proposed \textsc{Theory} (phase alignment plus the $\tau$ of Proposition~\ref{prop:tau}) and \textsc{Info} rules.
\end{itemize}
Each flight is split into six non-overlapping 10-minute windows, and each window is run with five noise seeds, giving 30 runs per policy. Runs are \emph{paired}: all policies see the same windows, the same initial offset, and the same random seed, and differ only in sensing choices. The particle filter uses $N_p=2000$ particles. We report horizontal RMSE over the whole window and over its last 5 minutes.

Policies are compared by the median of per-run error ratios, with paired Wilcoxon signed-rank tests. Because seeds within a window are correlated, we also report window-level tests (log-ratios averaged per window) and Holm-corrected $p$-values across the 12 main conditions. A run counts as diverged if its mean error over the last 50\,s exceeds 500\,m.

\section{Results}\label{sec:results}
\begin{figure}[t]\centering
\includegraphics[width=\columnwidth]{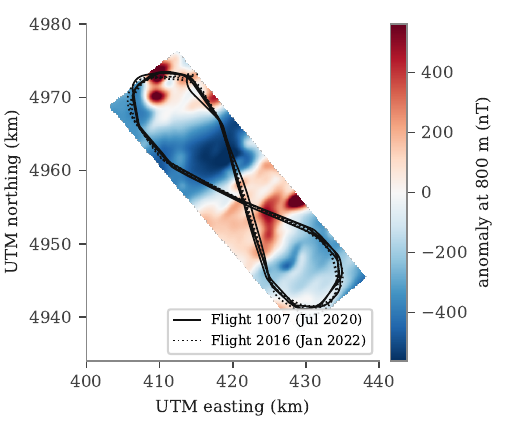}
\caption{Official Perth anomaly map at 800\,m (MagNav.jl compilation of the survey), with the two test tracks. Flight~2016 re-flies the Flight~1007 path about 18 months later.}
\label{fig:map}
\end{figure}

\subsection{How the rule behaves}
\begin{figure}[t]\centering
\includegraphics[width=\columnwidth]{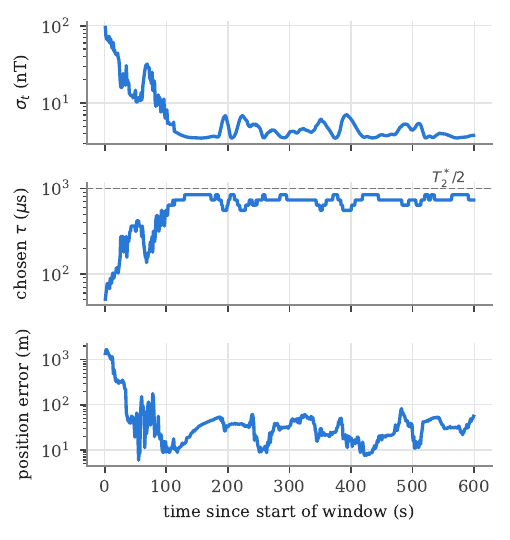}
\caption{One lost-start window (Flight~1007, official map, R4). Top: the filter's predicted-field spread $\sigma_t$. Middle: the interrogation time chosen by \textsc{Theory}, which moves from the ambiguity limit toward the decoherence limit $T_2^*/2$. Bottom: position error.}
\label{fig:trace}
\end{figure}

Fig.~\ref{fig:trace} traces one lost-start window under \textsc{Theory}:
\begin{itemize}
\item At the start, the predicted-field spread is about \trSdStart\,nT. The rule therefore chooses a short interrogation time (\trTauStart\,$\mu$s), which keeps the fringe unambiguous across the hypotheses.
\item As the filter localizes, the spread falls to about \trSdEnd\,nT, which is close to the map-error floor. The rule lengthens $\tau$ to about \trTauEnd\,$\mu$s, near the decoherence limit $T_2^*/2$ of Proposition~\ref{prop:tau}.
\item Position error falls from kilometres to about \trErrEnd\,m.
\end{itemize}
No fixed setting can follow this path: a short $\tau$ wastes sensitivity once the vehicle is localized, and a long $\tau$ aliases while it is lost.

\begin{table*}[t]\centering\caption{Real-flight results with the official map (map error 3.2--3.4\,nT). 30 paired runs per cell. ``Best'' is the oracle-chosen best of the fixed-$\tau$ settings and the ladder. The ratio is the median over runs of Theory RMSE divided by best-baseline RMSE (below 1 means Theory is better); $p$ is a paired Wilcoxon test on whole-window RMSE.}\label{tab:official}
\setlength{\tabcolsep}{4pt}
\begin{tabular}{llrrrcccrc}\toprule
Flight & Sensor class & \multicolumn{3}{c}{Mean RMSE (m)} & \multicolumn{2}{c}{Theory/best, median ratio} & Wins & $p$ & Diverged \\
\cmidrule(lr){3-5}\cmidrule(lr){6-7}
 & (best baseline) & Best & Theory & Info & whole & last 5 min & /30 & & best/Th./Info\\\midrule
1007 & R1 ensemble (Ladder) & 62 & 59 & 62 & 0.91 & 0.88 & 22 & 0.01 & 0/0/0\\
2016 & R1 ensemble (Ladder) & 74 & 69 & 65 & 0.92 & 0.99 & 18 & 0.08 & 0/0/0\\
\addlinespace
1007 & R3 low-SNR (Fixed\,100\,$\mu$s) & 299 & 217 & 253 & 0.80 & 0.68 & 22 & 0.002 & 0/0/1\\
2016 & R3 low-SNR (Fixed\,100\,$\mu$s) & 300 & 224 & 241 & 0.73 & 0.60 & 24 & 0.0002 & 3/0/0\\
\addlinespace
1007 & R4 lost start (Fixed\,20\,$\mu$s) & 687 & 805 & 645 & 0.55 & 0.21 & 28 & 5e-06 & 1/1/2\\
2016 & R4 lost start (Ladder) & 690 & 753 & 794 & 0.70 & 0.62 & 23 & 0.006 & 1/2/3\\
\bottomrule\end{tabular}\end{table*}

\subsection{Official map}
\begin{figure}[t]\centering
\includegraphics[width=\columnwidth]{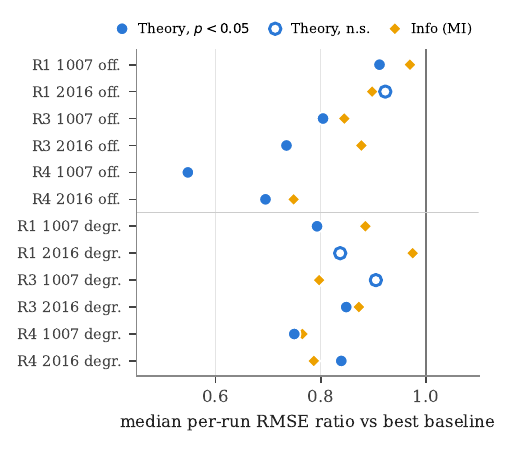}
\caption{Median per-run RMSE ratio against the best (hindsight-tuned) baseline in all 12 conditions. ``off.'' is the official map; ``degr.'' is the degraded map. Filled circles: \textsc{Theory} significant at $p<0.05$ (paired Wilcoxon, 30 runs, uncorrected; Holm-corrected counts are given in the text). Values below 1 favour belief-driven sensing.}
\label{fig:ratios}
\end{figure}

Table~\ref{tab:official} compares \textsc{Theory} with the best fixed-reference baseline (fixed $\tau$ or ladder) for each case, chosen with hindsight. Across all \nCond\ conditions (both maps, both flights, three sensor classes; Fig.~\ref{fig:ratios}):
\begin{itemize}
\item the median per-run error ratio lies between \ratioMin\ and \ratioMax;
\item it is significant at $p<0.05$ in \nCondSig\ of \nCond\ before correction, and in \holmSig\ of \nCond\ after Holm correction;
\item at the level of independent flight windows, \winWins\ of \winN\ windows favour \textsc{Theory} (one-sided sign test $p=2\times10^{-11}$).
\end{itemize}
The size of the gain depends strongly on the sensor class:
\begin{itemize}
\item \emph{Ensemble sensor (R1).} Small gains only: median ratios \rOneRatioA\ and \rOneRatioB\ ($p=0.01$ and \rOnePB). With a sensitive sensor and an accurate map, accuracy is limited by the map, and every reasonable sensing schedule ends up close to that floor.
\item \emph{Low-SNR sensor (R3).} Whole-window error is \rThreeWholeLo--\rThreeWholeHi\% lower, and error over the last 5 minutes is \rThreeLastLo--\rThreeLastHi\% lower, on both flights. This includes Flight~2016, flown 18 months after the map. \textsc{Theory} also had no divergences, whereas the best fixed setting diverged in 3 of 30 runs on Flight~2016.
\item \emph{Lost start (R4).} On Flight~1007 the median whole-window error is \rFourWholeA\% lower, and the median error over the last 5 minutes is \rFourLastA\% lower ($\rFourWinsA/30$ wins). The median final error is \rFourMedTheoryA\,m against \rFourMedBestA\,m. On Flight~2016 the reductions are \rFourWholeB\% and \rFourLastB\%. However, a few lost-start runs diverge under both the adaptive and the best baseline policies (\divTheoryMin--\divTheoryMax\ of 30 for \textsc{Theory}, \divBaseMin--\divBaseMax\ for the best baseline, across maps and flights). Because divergences dominate a mean, the \emph{mean} RMSE of \textsc{Theory} in R4 is not lower than the best baseline's, even though the typical run is far better. Divergence is a property of the map-matching filter under large initial uncertainty, which sensing choices alone do not remove (Sec.~\ref{sec:discussion}).
\end{itemize}

\subsection{What drives the gain: phase alignment versus interrogation time}\label{sec:ablation}
\textsc{Theory} changes two things relative to the fixed-reference baselines: it aligns the phase to the predicted field, and it adapts $\tau$. Table~\ref{tab:ablation} separates the two.
\begin{itemize}
\item \emph{Steady flight, low-SNR sensor (R3).} Phase alignment carries the gain. A fixed $\tau$ with aligned phase cuts error by \abBPthreeLo--\abBPthreeHi\% relative to the fixed-reference baseline, and \textsc{Theory} merely matches the best such setting (ratios \abRthreeA\ and \abRthreeB, $p\ge\abRthreePmin$). That best fixed $\tau$, however, was picked with hindsight from four candidates. \textsc{Theory} reaches the same accuracy with no tuning, which matters when the right fixed setting depends on the sensor, map and terrain.
\item \emph{Lost start (R4).} Interrogation-time adaptation is essential. Phase alignment alone does not help (ratios \abBPfourA\ and \abBPfourB\ against the fixed-reference baseline). $\tau$-adaptation alone cuts median error by \abFPfourLo--\abFPfourHi\%, and \textsc{Theory} beats the hindsight-best phase-aligned fixed $\tau$ by \abRfourLo--\abRfourHi\% ($p\le\abRfourPmax$). This is the regime that follows a loss of position under jamming.
\item \emph{Ensemble sensor (R1).} Both ingredients give small gains (\textsc{Theory} against the best phase-aligned fixed $\tau$: \abRoneA, \abRoneB).
\item \emph{Ladders.} Reading both quadratures (\textsc{Ladder2}) or aligning the ladder's phase (\textsc{Ladder}+BP) does not close the gap: \textsc{Theory}'s median ratio against them is \abLadMin--\abLadMax\ in all six cases.
\end{itemize}
One caution. In the lost start, the $\tau$\textsc{-only} variant has a lower \emph{mean} RMSE than \textsc{Theory} (\abFPmeanRA\ vs.\ \abThMeanRA\,m and \abFPmeanRB\ vs.\ \abThMeanRB\,m), though not a lower median. Means here are dominated by the 1--3 diverging runs out of 30. We tested whether aligning the phase to the \emph{dominant mode} of the predicted field, rather than its mean, would remove these failures. It did not. Mean RMSE improved on Flight~1007 (519 vs.\ 805\,m) but worsened on Flight~2016 (1224 vs.\ 753\,m), and medians were unchanged (ratios 0.99--1.08). The rare divergences therefore stem from the filter's handling of map ambiguity under large initial uncertainty, not from the phase-alignment choice.

\emph{Heuristic constants.} Among the three ambiguity-limited $\tau$ rules, \textsc{Theory} (implied $\kappa=1/\sqrt2$) gives the lowest error ratio in 6 of the 12 conditions, and \textsc{Heur}($\kappa{=}1$) and \textsc{Heur}($\kappa{=}0.5$) in three each. The closed form thus removes the need to tune $\kappa$ without giving up accuracy.

\begin{table}[t]\centering\caption{Ablation on the official map: median per-run RMSE ratio of \textsc{Theory} to stronger baselines (30 paired runs; $^{**}p<0.01$, $^{*}p<0.05$). BP = phase aligned to the predicted field. Last column: $\tau$ rule \emph{without} phase alignment against the best fixed-reference baseline.}\label{tab:ablation}
\setlength{\tabcolsep}{2.5pt}\begin{tabular}{llcccc}\toprule
Flt. & Class & best fixed-$\tau$+BP & Ladder2 & Ladder+BP & $\tau$-only\\\midrule
1007 & R1 ensemble & 0.95$^{*}$ & 0.92$^{**}$ & 0.94 & 0.94\\
2016 & R1 ensemble & 0.87$^{**}$ & 0.93 & 0.93$^{*}$ & 0.92$^{*}$\\
1007 & R3 low-SNR & 1.00 & 0.61$^{**}$ & 0.77$^{**}$ & 0.77$^{**}$\\
2016 & R3 low-SNR & 0.98 & 0.68$^{**}$ & 0.66$^{**}$ & 0.83$^{**}$\\
1007 & R4 lost start & 0.58$^{**}$ & 0.74$^{**}$ & 0.77$^{**}$ & 0.73$^{**}$\\
2016 & R4 lost start & 0.63$^{**}$ & 0.73$^{**}$ & 0.81$^{**}$ & 0.80$^{**}$\\
\bottomrule\end{tabular}\end{table}

\subsection{Which sensors benefit}
\begin{figure}[t]\centering
\includegraphics[width=\columnwidth]{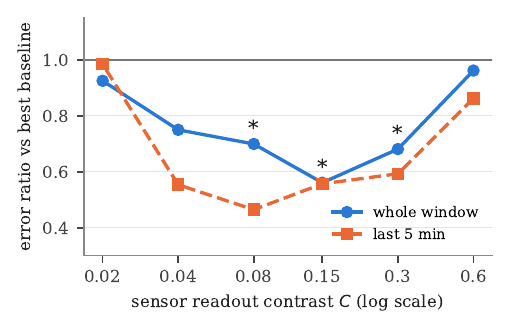}
\caption{Benefit of \textsc{Theory} as a function of sensor readout contrast (Flight~1007, official map). Error ratio against the best fixed or ladder baseline, chosen per contrast with hindsight. $*$: whole-window $p<0.05$.}
\label{fig:sweep}
\end{figure}

Fig.~\ref{fig:sweep} sweeps the readout contrast $C$ on Flight~1007, with all other settings fixed (18 paired runs per point). The benefit has an inverted-U shape:
\begin{itemize}
\item \emph{Mid-range sensors} ($C=0.08$--$0.3$): error is \swMidLo--\swMidHi\% lower over the whole window (all $p\le\swMidPmax$), and up to \swMidLastHi\% lower over the last 5 minutes.
\item \emph{Top-quality sensors} ($C=0.6$): the ratio is \swHighRatio\ ($p=\swHighP$). The sensor is already map-limited.
\item \emph{Very weak sensors} ($C=0.02$): the ratio is \swLowRatio\ ($p=\swLowP$). Each epoch carries too little information for scheduling to matter much.
\end{itemize}
Belief-driven sensing therefore pays off most for the compact, moderate-contrast sensors that low size-weight-and-power platforms such as small UAVs can carry.

\begin{table*}[t]\centering\caption{Stress test with the degraded survey-built map (map error about 20\,nT). Same protocol as Table~\ref{tab:official}.}\label{tab:survey}
\setlength{\tabcolsep}{4pt}
\begin{tabular}{llrrrcccrc}\toprule
Flight & Sensor class & \multicolumn{3}{c}{Mean RMSE (m)} & \multicolumn{2}{c}{Theory/best, median ratio} & Wins & $p$ & Diverged \\
\cmidrule(lr){3-5}\cmidrule(lr){6-7}
 & (best baseline) & Best & Theory & Info & whole & last 5 min & /30 & & best/Th./Info\\\midrule
1007 & R1 ensemble (Ladder) & 110 & 93 & 117 & 0.79 & 0.78 & 27 & 0.0002 & 0/0/0\\
2016 & R1 ensemble (Fixed\,20\,$\mu$s) & 176 & 294 & 199 & 0.84 & 0.55 & 19 & 0.8 & 0/1/0\\
\addlinespace
1007 & R3 low-SNR (Fixed\,100\,$\mu$s) & 292 & 245 & 305 & 0.90 & 0.89 & 18 & 0.05 & 1/0/1\\
2016 & R3 low-SNR (Fixed\,100\,$\mu$s) & 320 & 250 & 248 & 0.85 & 0.79 & 19 & 0.03 & 4/0/0\\
\addlinespace
1007 & R4 lost start (Ladder) & 845 & 622 & 538 & 0.75 & 0.83 & 27 & 2e-05 & 2/1/1\\
2016 & R4 lost start (Ladder) & 700 & 603 & 579 & 0.84 & 0.94 & 20 & 0.03 & 1/1/2\\
\bottomrule\end{tabular}\end{table*}

\subsection{Degraded map}
With the 20\,nT survey-built map (Table~\ref{tab:survey}), \textsc{Theory} reduces median error by \degrLo--\degrHi\% in all six conditions, and the reduction is significant in \degrSig\ of six. Because the predictive spread includes the map error, the rule automatically shortens $\tau$ when the map is poor, so it does not over-commit to a long, ambiguity-prone interrogation time.

\subsection{Joint route and sensor control}
\emph{Setup.} The routes are counterfactual: they leave the flown tracks, where no measured field exists. The true field along them is therefore the official map plus a spatially correlated map-error field, with standard deviation 3.4\,nT (as measured) and correlation length $\approx$700\,m, drawn independently for each mission. Each mission crosses the 40$\times$12\,km survey from one end to the other, about 34\,km, at 60\,m/s, with randomized start and goal lateral offsets. A navigation-grade INS drift is simulated. The filter and sensor models are those of Sec.~\ref{sec:problem}, and the fixed sensor uses $\tau=100\,\mu$s. Each cell has 30 paired missions.

\begin{table}[t]\centering\caption{Joint route and sensor control (official Perth map with correlated 3.4\,nT map error; 30 paired missions per cell). Entries: median per-mission RMSE ratio against straight route with fixed sensor; ** $p<0.01$, * $p<0.05$ (paired Wilcoxon). Last row: median flight-time overhead.}\label{tab:plan}
\setlength{\tabcolsep}{3pt}\begin{tabular}{lcccccc}\toprule
Route & \multicolumn{2}{c}{Straight} & \multicolumn{2}{c}{Info-light ($\pm25^\circ$)} & \multicolumn{2}{c}{Info ($\pm50^\circ$)}\\
\cmidrule(lr){2-3}\cmidrule(lr){4-5}\cmidrule(lr){6-7}
Sensor & fixed & belief & fixed & belief & fixed & belief\\\midrule
R1 ensemble & 1 & 0.76** & 0.89** & 0.73** & 0.84** & 0.67**\\
Mid ($C{=}0.15$) & 1 & 0.66** & 0.96 & 0.57** & 0.94* & 0.54**\\
R3 low-SNR & 1 & 0.86 & 0.99 & 0.73** & 1.00 & 0.77**\\
R4 lost start & 1 & 0.74** & 1.01 & 0.71** & 1.02 & 0.67**\\
\midrule Time & \multicolumn{2}{c}{+0\%} & \multicolumn{2}{c}{+9\%} & \multicolumn{2}{c}{+29\%}\\
\bottomrule\end{tabular}\end{table}

\begin{figure}[t]\centering
\includegraphics[width=\columnwidth]{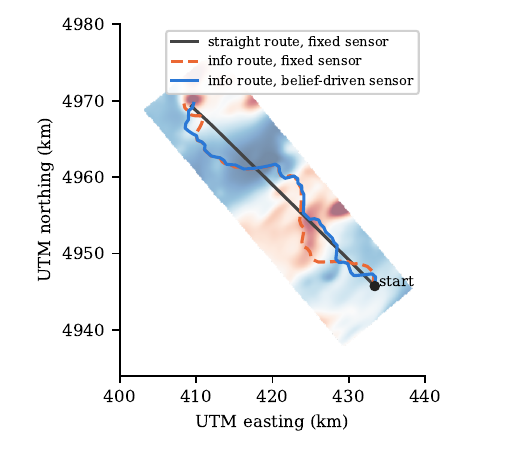}
\caption{Example mission (mid-contrast sensor, same seed). The information-aware routes detour over gradient-rich terrain, where map-matching information is highest.}
\label{fig:paths}
\end{figure}

\emph{Results} (Table~\ref{tab:plan}, Fig.~\ref{fig:paths}). Because the fixed-reference sensor is confounded by phase misalignment (Sec.~\ref{sec:ablation}), we also ran a phase-aligned fixed sensor ($\tau=100\,\mu$s+BP) on both route types.
\begin{enumerate}
\item \emph{Joint control is best.} Information-aware routing with belief-driven sensing lowers median mission RMSE by \planJointLo--\planJointHi\% against a straight route with a fixed-reference sensor. Against the stronger baseline, a straight route with a phase-aligned fixed sensor, it still lowers error by \plJointBPLo--\plJointBPHi\% in all four classes ($p\le\plJointBPPmax$).
\item \emph{Routing helps a phase-aligned sensor too.} With a phase-aligned fixed sensor, information-aware routing removes \plBProuteLo--\plBProuteHi\% of the error (significant in \plBProuteSig\ of 4 classes). Adding $\tau$-adaptation on top removes a further \plMidJointInfo\% for the mid-contrast sensor and \plLostJointInfo\% after a lost start, both significant. For the ensemble and low-SNR classes the further gain is small and not significant.
\item \emph{Most of the gain is cheap.} The light planner ($\pm25^\circ$) achieves \planLightLo--\planLightHi\% lower error with belief-driven sensing, for a median \planOvLight\% longer flight, against \planOvFull\% for the full planner.
\end{enumerate}
At most \planDivMax\ of 30 missions diverged in any cell.

\subsection{Hardware check on a superconducting qubit}\label{sec:hw}
Proposition~\ref{prop:tau} rests on two assumptions: the Ramsey model of Eq.~\eqref{eq:ramsey}, and visibility weighting by the predictive spread. We tested both on IBM's \texttt{ibm\_kingston} processor (job \texttt{\hwJob}) with single-qubit Ramsey circuits $\sqrt{X}$--delay$(\tau)$--$R_z(2\pi f\tau+\pi/2)$--$\sqrt{X}$, so that the qubit genuinely decoheres during the wait. The ``field'' is a detuning $f$ drawn from a Gaussian predictive belief with spread $\sigma_f\in\{1,5,20\}$\,kHz, which is equivalent to about 36, 180 and 710\,nT at $\gamma=28$\,Hz/nT. The circuits cover ten values of $\tau$ from 2 to 250\,$\mu$s, with 16 draws per setting, 512 shots each, and 40 calibration circuits, 520 circuits in all. The whole experiment ran as a single job: about 3\,min of wall-clock time from submission to completion, consuming 1.49 IBM resource units under the free Open Plan.

From the measured shots we emulate a per-epoch budget of 10\,ms with 20\,$\mu$s overhead per shot. The bright-shot counts are then combined with the prior in a Bayesian estimate, and we report its normalized mean-squared error (nMSE). Calibration gives a contrast of \hwAzero\ and an \emph{exponential} decay with $T_2^*=\hwTtwo\,\mu$s (fitted stretch exponent \hwBeta), rather than the Gaussian decay assumed in Eq.~\eqref{eq:ramsey}. We therefore evaluate the rule with the measured decay law, maximizing Eq.~\eqref{eq:score} numerically.

\begin{figure}[t]\centering
\includegraphics[width=\columnwidth]{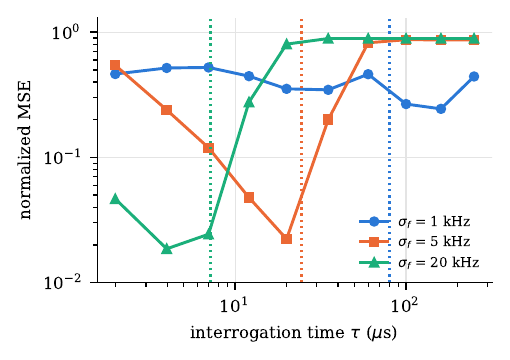}
\caption{Hardware check on \texttt{ibm\_kingston}: Bayesian estimation error against interrogation time for three predictive spreads. Dotted lines mark the $\tau$ predicted by the visibility-weighted rule (Proposition~\ref{prop:tau} with the measured decay law), computed from the calibration circuits alone, independently of the estimation results.}
\label{fig:hw}
\end{figure}

Fig.~\ref{fig:hw} shows the trade-off the paper is built on:
\begin{itemize}
\item every predictive spread has its own optimal $\tau$, and a $\tau$ that is optimal for one spread is far from optimal for another;
\item the predicted $\tau$ lands at or near each empirical optimum, and its error is within a factor \hwAdaptWorst\ of the best grid value at every spread;
\item every \emph{fixed} $\tau$ is worse than the per-spread optimum by at least a factor \hwFixedBest\ in some regime (best case $\tau\approx\hwFixedBestTau\,\mu$s), and by up to \hwFixedWorst$\times$.
\end{itemize}
The rule is slightly long at the largest spread. There the Bayesian estimator still extracts some information from aliased fringes, which the Gaussian visibility argument ignores. This matches the conservative bias seen in simulation.

\subsection{Information-gain rule}
\textsc{Info} beat \textsc{Theory} in only \infoBetterCount\ of 12 conditions, and it costs about eight times more computation per epoch. Under the beliefs encountered here, the closed-form Gaussian rule already captures most of the available benefit. Mutual-information planning may still matter for joint sensor and trajectory control, where the action space is larger.

\section{Discussion and Limitations}\label{sec:discussion}
\emph{Scope of the claim.} Belief-driven sensing does not make a map-limited system better. When the sensor is excellent and the map accurate, every reasonable schedule converges to the map-error floor, and the gain is a few percent. Its value lies in two places:
\begin{itemize}
\item \emph{sensor-limited operation}: moderate- and low-contrast sensors, short sensing budgets, small platforms;
\item \emph{transients}: a lost start, or recovery after a gap.
\end{itemize}
In both, the right operating point changes by more than an order of magnitude during a flight (Fig.~\ref{fig:trace}).

\emph{Implications for small UAS.} The regimes where belief-driven sensing pays off are those of small, low-SWaP drones.
\begin{itemize}
\item \emph{Sensor quality.} Compact, low-power quantum magnetometers on Group~1--3 aircraft have weak readout: the low-SNR and mid-contrast classes, where we see 20--44\% lower error.
\item \emph{Loss of position.} Jamming produces sudden losses of position, the lost-start regime. There interrogation-time adaptation, not phase alignment, cuts median error by \abRfourLo--\abRfourHi\%.
\item \emph{Cost and integration.} The rule is closed form, costs microseconds per epoch on an embedded processor, and requires no new hardware, only control of the sensor's timing. It can therefore be retrofitted to any Ramsey-type magnetometer and to existing particle-filter MagNav stacks.
\item \emph{Routing.} For drones with route flexibility, such as survey or reconnaissance missions, information-aware routing adds a further, cheap improvement: most of the gain comes for a 9\% longer flight.
\end{itemize}
Conversely, for large platforms with high-grade sensors and accurate maps, the method offers little. The method is therefore best suited to low-cost platforms, where sensor quality, not the map, limits accuracy.

\emph{Limitations.}
\begin{enumerate}
\item \emph{The Ramsey sensor is modelled.} In the sensor-control study the trajectories, inertial drift, measured anomaly fields and map errors are real. The Ramsey shots are simulated from Eq.~\eqref{eq:ramsey}, because the dataset records continuous optically pumped magnetometers. The sensor classes are generic rather than specific devices. A single-qubit hardware check (Sec.~\ref{sec:hw}) supports the measurement model and the visibility-weighted rule. It does not replace a navigation-grade magnetometer, whose decoherence time and readout differ.
\item \emph{Map level.} We remove a constant level offset between map and measurements for each flight, estimated from the full track. The offsets are small ($-1.8$ and $+2.2$\,nT, below $\sigma_m$), but a deployed system would estimate them online as an extra filter state.
\item \emph{Map-error model.} The likelihood treats map error as white, whereas the measured errors are correlated over about 10\,s. This model mismatch affects all policies equally, and likely contributes to the rare lost-start divergences.
\item \emph{Divergence.} After a lost start, 1--3 of 30 runs diverge under the adaptive policies, as do 0--2 under the strongest baselines. Mode-aware phase alignment did not remove them (Sec.~\ref{sec:ablation}), so the cause lies in the filter, not the sensor schedule. Robust likelihoods, multi-hypothesis re-initialization, or integrity monitoring are needed. They are complementary to sensor control.
\item \emph{Route planning uses counterfactual fields.} Planned routes leave the flown tracks, so their true field is the real map plus a simulated correlated error field, and their INS drift is simulated. The planner is greedy over a 10\,s horizon. Longer-horizon optimal control~\cite{kang2026optimal} or global entropy maps~\cite{penumarti2024global} could replace it.
\item \emph{Platform and altitude.} The data come from a crewed survey aircraft at 800\,m over one area. Small drones fly lower, where anomaly gradients are stronger but platform interference from motors and electronics is larger. Validating on sUAS flight data, with upward- or downward-continued maps, is the necessary next step.
\end{enumerate}

\emph{Relation to adaptive phase estimation.} Bayesian adaptive magnetometry chooses sensing times from a posterior built \emph{by the sensor itself} to estimate a static field~\cite{said2011,bonato2016,ma2025adaptive}. Here the prior comes from the navigator: the map plus the inertial dead-reckoning. The objective is position, and the prior is refreshed every epoch as the vehicle moves. Proposition~\ref{prop:tau} makes the coupling explicit: the navigator's uncertainty sets the sensor's ambiguity limit.

\section{Conclusion}
We closed the loop between a map-matching navigation filter and the operating point of a Ramsey quantum magnetometer. The filter's predictive field belief aligns the readout phase and sets the interrogation time at every epoch, through a closed-form rule that moves between an ambiguity limit and a decoherence limit.

On real airborne data (real tracks, inertial drift, measured fields, and a map tested 18 months after it was flown):
\begin{itemize}
\item \winWins\ of \winN\ independent flight windows favoured belief-driven sensing over hindsight-tuned fixed-reference settings;
\item error fell by \rThreeWholeLo--\rThreeWholeHi\% for low-SNR sensors and \swMidLo--\swMidHi\% for mid-contrast sensors;
\item after a lost start, interrogation-time adaptation cut median error by \abRfourLo--\abRfourHi\% even against hindsight-tuned phase-aligned settings;
\item gains were small for high-quality sensors with accurate maps.
\end{itemize}
A superconducting-qubit experiment confirmed the rule's interrogation time to within $1.3\times$ of the optimum. Joint route and sensor control lowered mission error by \plJointBPLo--\plJointBPHi\% against a phase-aligned fixed sensor on a straight route.

For small, low-SWaP drones, which carry the weakest sensors and suffer the most abrupt losses of position under jamming, this is a software-only improvement that costs microseconds per epoch. Next steps are flight tests on Group~2--3 sUAS with compact quantum magnetometers, online map-level estimation, divergence-robust filtering after a lost start, and longer-horizon planning.

\section*{Acknowledgment and Data Statement}
DAF-MIT AIA Open Flight Data for Magnetic Navigation Research provided by the United States Air Force pursuant to Cooperative Agreement Number FA8750-19-2-1000 -- November 2026 (flights recorded 2020--2022)~\cite{dafmitdata,gnadt2023signal}. The Perth anomaly map is the MagNav.jl compilation of the Sander Geophysics Ltd.\ survey. We acknowledge the use of IBM Quantum services for this work. The views expressed are those of the authors, and do not reflect the official policy or position of IBM, the IBM Quantum team, or the U.S.\ Air Force. Generative AI tools (Anthropic Claude) assisted with code, analysis and drafting; the authors reviewed all content and take full responsibility for it. Code will be made available; the flight data must be obtained from the original source under the Air Force Acceptable Use Agreement.

\bibliographystyle{IEEEtran}
\bibliography{refs}
\end{document}